\documentclass{article}

    \usepackage[preprint]{neurips_2024}

\usepackage[utf8]{inputenc} 
\usepackage[T1]{fontenc}    
\usepackage{hyperref}       
\usepackage{url}            
\usepackage{booktabs}       
\usepackage{amsfonts}       
\usepackage{nicefrac}       
\usepackage{microtype}      
\usepackage{xcolor}         

\usepackage{amsmath}
\usepackage{graphicx}
\usepackage{subcaption}

\usepackage{natbib}
\title{\texttt{gzip} Predicts Data-dependent Scaling Laws}

\author{
  Rohan Pandey \\
  Reworkd \\
  San Francisco, CA \\
  \texttt{rohan@reworkd.ai} \\
}

\begin{document}

\maketitle

\begin{abstract}
    Past work has established scaling laws that predict the performance of a neural language model (LM) as a function of its parameter count and the number of tokens it's trained on, enabling optimal allocation of a fixed compute budget. Are these scaling laws agnostic to training data as some prior work suggests? 
    We generate training datasets of varying complexities by modulating the syntactic properties of a PCFG, finding that 1) scaling laws are sensitive to differences in data complexity and that 2) \texttt{gzip}, a compression algorithm, is an effective predictor of how data complexity impacts scaling properties. We propose a new data-dependent scaling law for LM's that accounts for the training data's \texttt{gzip}-compressibility; its compute-optimal frontier increases in dataset size preference (over parameter count preference) as training data becomes harder to compress.
\end{abstract}

\section{Introduction}

\label{intro}

A neural network's performance generally increases as more compute is allocated for training. When scaling compute, one must decide whether to increase a model's parameter count or increase the dataset's size—these must trade-off within a fixed compute budget. Scaling laws can tell us what specific allocation (i.e. parameters v.s. data) will maximize performance given a compute budget. Much work has explored scaling laws for neural LM's \citep{kaplan2020scaling}, generally concluding that parameter \& training token count should be scaled 1-to-1 \citep{hoffmann2022training}.

However, most prior work on scaling laws for LM's have been estimated from transformers trained on scraped web text. Of course, this is quite a specific data distribution, so we may naturally ask whether the scaling laws extrapolated from such web text datasets generalize to other distributions. Furthermore, it is generally understood that the art of training data mixture is the `secret sauce' that enables frontier industry labs to continually deliver state-of-the-art LLMs \citep{penedo2024fineweb, penedo2023refinedweb, xie2024doremi}.
Considering that improving data quality can significantly raise LM performance \citep{gunasekar2023textbooks} and that scaling laws in reinforcement learning have been shown to scale with game difficulty \citep{jones2021scaling}, we may hypothesize that current LM scaling laws (e.g. Chinchilla \citep{hoffmann2022training}) are individual web-text-specific cases of a more general scaling law conditioned on properties of the training data.

Then, what properties of a training dataset of token sequences are neural scaling laws sensitive to? In other words, what can we measure about our data to more accurately predict the optimal compute allocation for training? Furthermore, is data-dependence of scaling laws only of theoretical interest, or can laws be considerably different for real-world datasets?

In order to study these questions, we seek a textual data setting where we can intuitively control its complexity as well as open avenues for information-theoretic understandings of why scaling laws are data-dependent. We therefore settle on probabilistic context-free grammars (PCFG) \citep{chomsky1956three} which are relatively naturalistic (can model natural language, code, etc.), controllable in syntactic complexity, and follow some well-understood information-theoretic principles \citep{chi-1999-statistical}.

We generate 6 datasets of varying complexity by modulating syntactic properties of a PCFG. For each of these datasets, we train LMs of 6 different sizes (4.4M to 1.4B parameters) and record results at 6 different train step counts (100K to 100M tokens). We then fit a scaling law for each dataset \citep{muennighoff2024scaling}, finding meaningful shifts in the law's parameters as syntactic complexity increases. Following prior work on entropy of formal grammars \citep{arora2022estimating}, we use median compressibility of each token sequence in a dataset as a measure of complexity that is straightforward to compute with \texttt{gzip} \citep{gailly1992gnu}.

We find that \textbf{as the training data becomes less compressible} (more complex), the scaling law's compute-optimal frontier gradually \textbf{increases its preference for dataset size} over parameter count. We then measure the compressibility of real-world code \& natural language datasets, showing that the former is considerably more compressible and thus subject to a predictably different scaling law. With some napkin math, we estimate that we could reach the performance of StarCoder \citep{li2023starcoder} with \textbf{24\% fewer FLOPs} (\$278,000 in H100 hours) since our compute-optimal scaling adjusts for data complexity.

\section{Related Work}

\subsection{Scaling Laws}
Early work established that a neural network's test error is a power law of training dataset size \citep{cortes1993learning}, model parameter count \citep{rosenfeld2019constructive}, and that the relationship holds over many orders of magnitude \citep{hestness2017deep}.
\citet{kaplan2020scaling} applied scaling laws to transformer-based language models and identified a compute-optimal frontier along which parameter \& dataset size should be scaled.
\citet{hoffmann2022training} propose the Chinchilla scaling laws, finding that \citet{kaplan2020scaling} and \citet{rae2021scaling} overparameterized their models and that the compute-optimal frontier requires parameter \& dataset size to be scaled equally (rather than parameters scaling at 3x the rate of data).

\citet{sorscher2022beyond} find that we can reach exponential (rather than power law) scaling on dataset size by pruning out redundant examples that do not provide much information to learn from.
\citet{liu2023neural} identify a mechanistic explanation of why scaling follows a power law on model width.

\citet{aghajanyan2023scaling} extend the Chinchilla scaling laws to a variety of modalities such as speech, image-text, and code.
\citet{caballero2023broken} propose a novel functional form for scaling laws that better models complex non-monotonic behavior and also apply it to several modalities including code.
Both these works and \citet{hoffmann2022training} (in Appendix C) find that \textbf{scaling behavior for code is different than for natural language}—code is easier to learn and its compute-optimal frontier prefers parameters slightly over data. 
\citet{bi2024deepseek} cursorily investigate scaling laws across datasets of different qualities, finding that cleaner \& \textbf{higher quality data results in the ``model scaling exponent increasing''} because ``high-quality data usually implies logical clarity and less predictive difficulty after sufficient training''.
However, none of these works identify an \textbf{underlying general principle that explains why scaling behavior varies} across data modalities \& complexities (or even just between code and natural language).

Meanwhile, \citet{jones2021scaling} explores scaling laws in the context of board games, finding that ``the compute required for a desired level of performance can be calculated directly from the board size''. If scaling laws smoothly scale with data complexity in the board game setting, do they also smoothly scale with textual data complexity? And how might we measure a text dataset's complexity?

\subsection{Syntax \& Information Theory}

\label{syntax_info_theory}

A long history of work in linguistics has sought to apply information-theoretic measures to natural language \citep{shannon1951prediction, cherry1953toward, harris1991theory, piantadosi2011word}.
Specifically, entropy (a fairly abstract information-theoretic measure) can be operationalized in a number of concrete ways to measure the informational complexity of a linguistic distribution \citep{arora2022estimating}.
One way to determine the `goodness' of an entropy measure over sequences is whether it grows as the syntactic complexity of a language increases (e.g. more production rules, more non-terminals, longer right-hand sides).
While entropy is straightforward to compute in closed-form on simple formal languages (i.e. Type 3 Chomskyan grammars, generated by Finite State Automata) \citep{grenander1967syntax, sanchez-etal-2018-derivational}, for Context-free (Type 2) grammars, we must estimate entropy from a set of generated samples \citep{chi-1999-statistical, corazza2007probabilistic}.

Leveraging the widely-recognized relationship between entropy and compression \citep{shannon1948mathematical, huffman1952method, ziv1977universal}, we can use \texttt{gzip} \citep{gailly1992gnu}, a lossless compression utility that implements DEFLATE \citep{deutsch1996deflate} (a combination of Huffman coding \& the Lempel-Ziv algorithm) to estimate the entropy of sampled linguistic sequences.

\citet{deletang2023language} adapt language models as lossless compressors of token sequences and provide new insights on scaling laws from this compression perspective, finding that beyond a critical model size, the compression rate (accounting for parameter count) reverses its improvement. In a somewhat similar vein, \citet{jiang-etal-2023-low} use \texttt{gzip} with k-Nearest-Neighbors to beat neural text embeddings.
Importantly, they both brought to attention the effectiveness of compression algorithms for heuristically measuring the structure learned by language models \citep{huang2024compression}.
Another line of recent work investigates the learnability of different languages by LMs, finding that natural languages vary in their learnability \citep{cotterell-etal-2018-languages} and that LMs struggle to model syntactically `impossible' languages \citep{kallini2024mission}.

\section{Modulating Data Complexity via Syntactic Properties of a PCFG}

Probabilistic Context-Free Grammars (PCFGs) are a fundamental tool in computational linguistics for modeling the syntax of natural languages. A PCFG extends the concept of a standard Context-Free Grammar (CFG) by associating probabilities with its production rules, enabling the representation of language ambiguity and variability in a quantifiable manner. These grammars generate trees where each node represents a syntactic category, and the edges represent production rules applied to generate sentences. When generating sentences from a PCFG, we probabilistically sample sequences of production rules to apply until all leaves of the tree are terminals (actual vocab tokens).

We can control the syntactic properties of a PCFG to naturalistically modulate the complexity of a textual dataset that we sample from its generated sentences. Specifically, our PCFG creation function accepts arguments for the number of terminals, number of non-terminals, the maximum length of the right-hand side of a production rule, and the maximum number of production rules allowed for any non-terminal (i.e. if this is 1, a given non-terminal will \textit{always} lead to the same right-hand side). Intuitively, as each of these values increase, the language's syntactic complexity also increases.

To create a PCFG from the above arguments, for each non-terminal, we randomly choose its number of productions (RHS options), each of those production's length, instantiate a production rule by randomly sampling from the terminals \& non-terminals, and assign it a probability normalized by the total RHS options for the non-terminal. We then collect all the generated production rules for all the non-terminals and instantiate a grammar using the PCFG package \citep{Breydo2021thomasbreydo} built on NLTK \citep{bird-loper-2004-nltk}.

We can then use the grammar that we have randomly (within given constraints) created to probabilistically sample sentences to construct a dataset of token sequences. To make it easier later to compare training across grammars that produce sentences of different average lengths, we decide to sample sentences into documents of the same token count (our LM's context length). We sample sentences from our grammar until we've filled up the context length, and if we overflow, we simply truncate the sequence. 

Sentences are composed of terminals that are just integers and can thus be treated as token IDs for our LM; we concatenate sentences with the unused integer 0, effectively corresponding to a period in natural language. To clarify, we do not generate strings that `look' like natural language and must be tokenized—the PCFG generates sequences of token IDs themselves. Now, we can generate 6 datasets of token sequences with varying complexities from just 6 sets of initial grammatical constraints.

\subsection{\texttt{gzip}-compressibility Measures Syntactic Complexity}

\label{gzip_measures}

Now we must choose a metric to estimate the complexity of our datasets. While for our PCFG-generated data, we could use a trivial function of its syntactic properties (e.g. number of non-terminals plus median production rule length), we would only be able to apply this metric to datasets where we precisely know the grammar that generated the data. Instead, we need a measure of complexity that we can apply to real-world token sequence datasets so that we can see whether our data-sensitive scaling results on PCFGs generalize to real-world datasets. While one might try to use grammar induction to identify the underlying syntax of a real-world dataset and then compute complexity from its syntactic properties, this is computationally intensive and difficult with noisy web-text data.

Therefore, as explained in Sec. \ref{syntax_info_theory}, we choose to use a compression algorithm, \texttt{gzip}, to estimate the complexity of a dataset by virtue of considerable theoretical work establishing $\text{compressibility}^{-1} \propto$ entropy \citep{shannon1948mathematical} and entropy $\propto$ syntactic complexity \citep{chi-1999-statistical}. Specifically, for each token sequence in a sample of 1000 from the dataset, we apply \texttt{gzip} and compute the ratio of the size (in bytes) of the compressed data to the original data. We then compute the median and standard deviation in compressibility, confirming that grammars with higher syntactic complexity result in datasets that are more difficult to compress.

\begin{figure}
    \centering
    \includegraphics[width=0.7\textwidth]{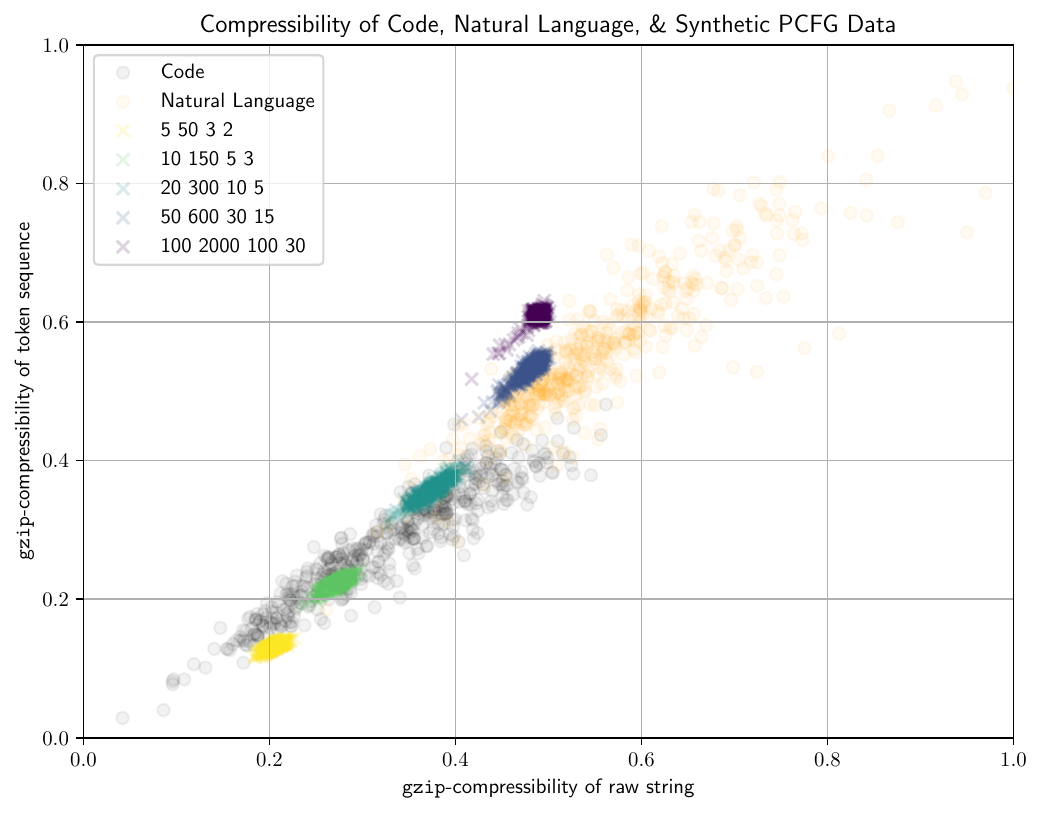}
    \caption{Plot of different linguistic distributions and their compressibility by \texttt{gzip} on their raw string \& their token sequence. The large grey and orange clusters are real-world code and natural language, respectively. The smaller clusters are sampled from PCFG's of increasingly complex syntax (with specific values for properties provided in legend). Observe that 1) natural language is more difficult to compress than code, and 2) as our PCFGs' syntactic complexity increases, the token sequences become less compressible. PCFG data is more tightly clustered due to length normalization.}
    \label{fig:compressibility}
\end{figure}

In Tab. \ref{tab:parameters}, we list the syntactic parameters for each grammar and the compression ratio we measured of token sequences sampled from the grammar. Observe that as non-terminals (grammatical categories), terminals (tokens), right-hand side options, and right-hand side length increase, the \texttt{gzip}-compressibility also increases (i.e. it becomes harder to compress). We plot these datasets alongside natural language \& code in Fig. \ref{fig:compressibility}, showing how some PCFG datasets are more similar in complexity to code (the ones that are easier to compress) while others are more similar to natural language.

\begin{table}[ht]
\centering
\caption{Compressibility of each dataset and the syntactic properties of its PCFG}
\begin{tabular}{cccccc}
\toprule
\texttt{gzip} Compressibility & Non-terminals & Terminals & RHS Options & RHS Length \\
\midrule
0.11 & 3 & 20 & 2 & 2 \\
0.22 & 10 & 150 & 5 & 3 \\
0.35 & 20 & 300 & 10 & 5 \\
0.42 & 30 & 400 & 10 & 8 \\
0.51 & 50 & 500 & 20 & 15 \\
0.61 & 100 & 2000 & 100 & 30 \\
\bottomrule
\end{tabular}
\label{tab:parameters}
\end{table}

\section{Are Scaling Laws Sensitive to Data Complexity?}

To identify the scaling law for a dataset, we train a set of models of varying sizes (4.2M, 8.8M, 20.3M, 59.0M, 275.3M, 1.4B parameters; architectural specifics in Tab. \ref{tab:architecture}) on varying size subsets of the data (100K, 1M, 5M, 20M, 50M, 100M tokens) and fit a power law on the resulting final losses of all training runs. Most experiments were run on a cluster of 4 Nvidia A100's with 80 GB VRAM each using PyTorch FSDP \citep{paszke2019pytorch, zhao2023pytorch}. We use a batch size of 32, single epoch training, the $AdamW$ optimizer \citep{loshchilov2017decoupled}, and a learning rate starting from $5\mathrm{e}{-5}$ and cosine decayed over the number of train steps \citep{loshchilov2016sgdr}.

\begin{figure}[ht]
    \centering
    
    \begin{subfigure}{0.32\textwidth}
        \includegraphics[width=\linewidth]{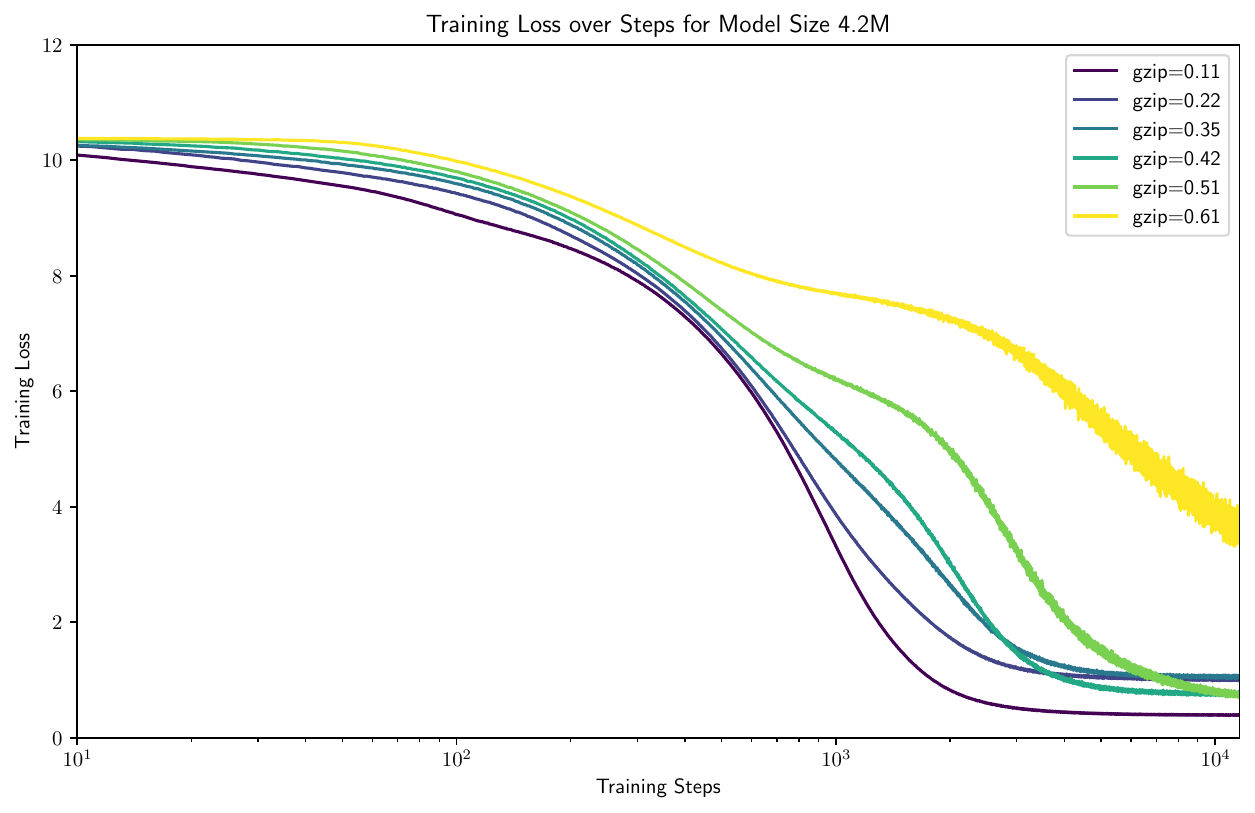}
        \caption{4.2M}
    \end{subfigure}
    \hfill
    \begin{subfigure}{0.32\textwidth}
        \includegraphics[width=\linewidth]{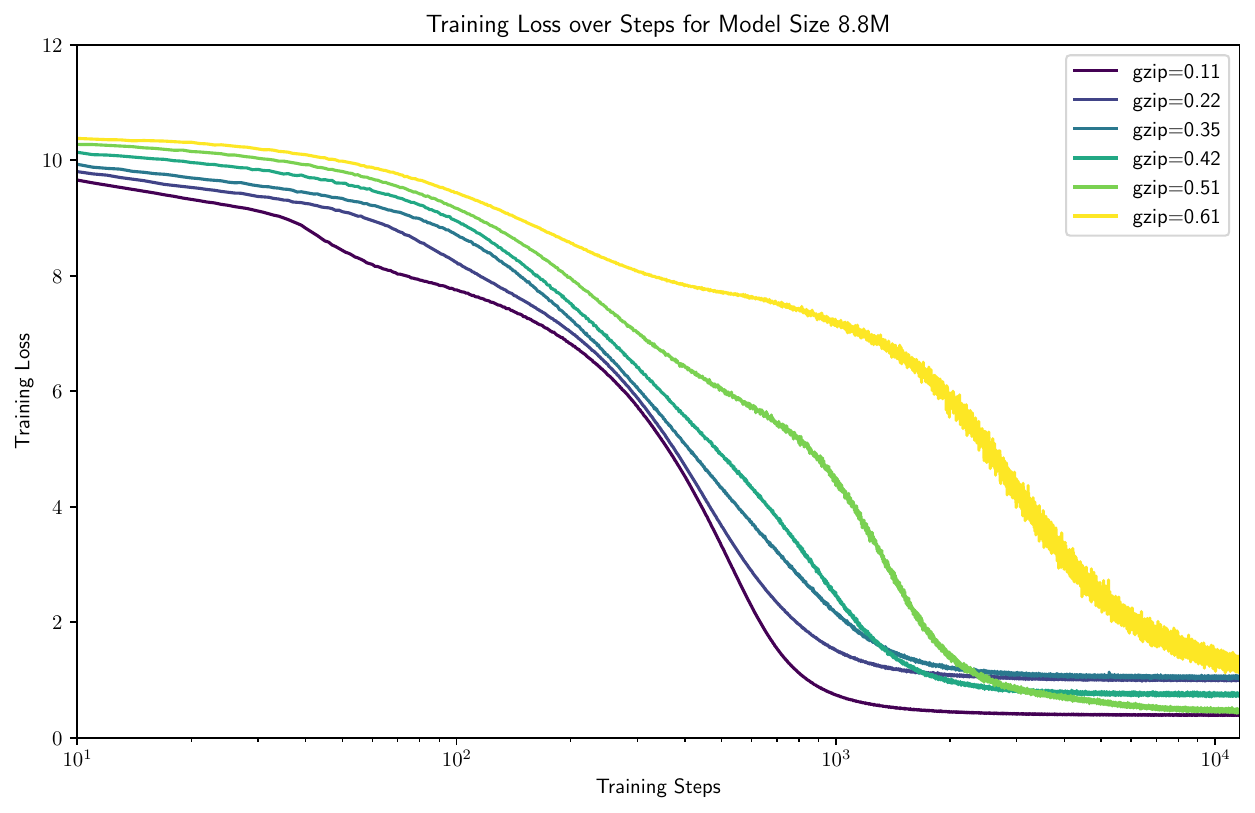}
        \caption{8.8M}
    \end{subfigure}
    \hfill
    \begin{subfigure}{0.32\textwidth}
        \includegraphics[width=\linewidth]{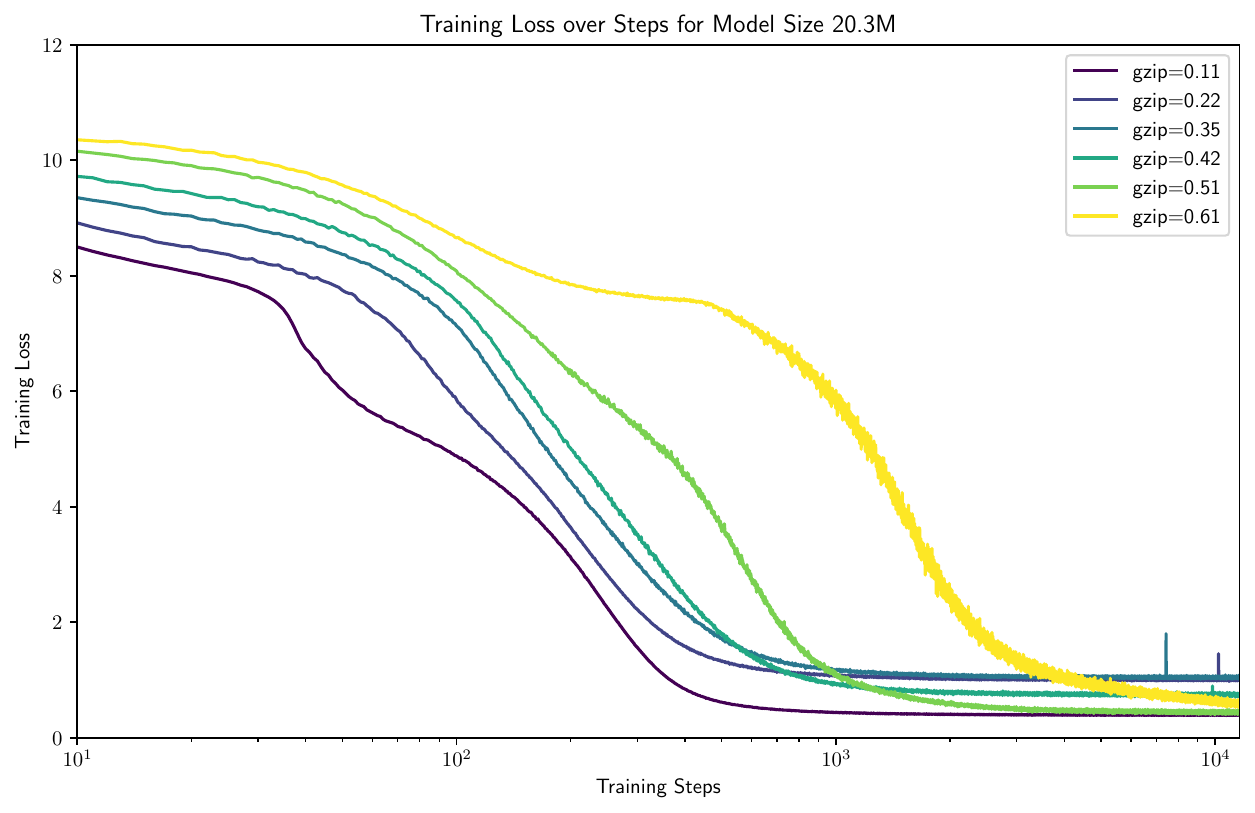}
        \caption{20.3M}
    \end{subfigure}

    \begin{subfigure}{0.32\textwidth}
        \includegraphics[width=\linewidth]{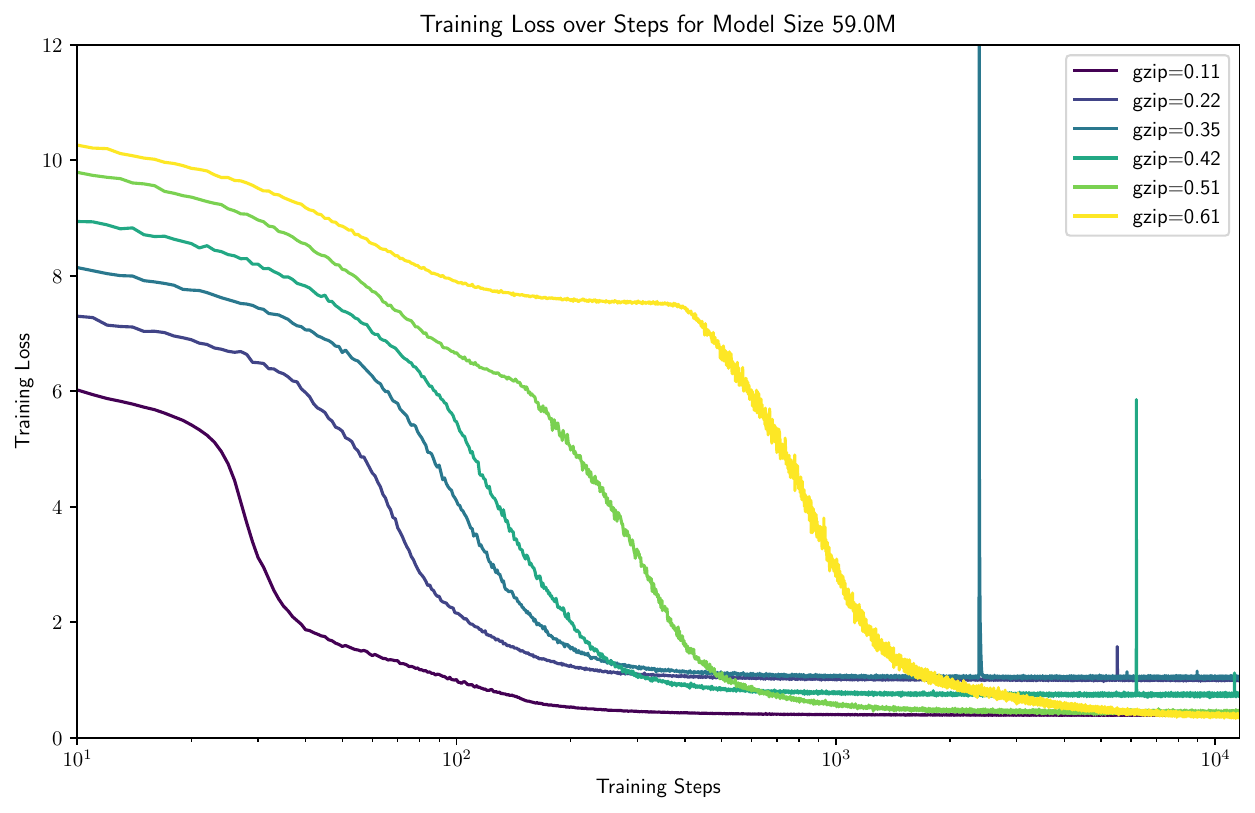}
        \caption{59.0M}
    \end{subfigure}
    \hfill
    \begin{subfigure}{0.32\textwidth}
        \includegraphics[width=\linewidth]{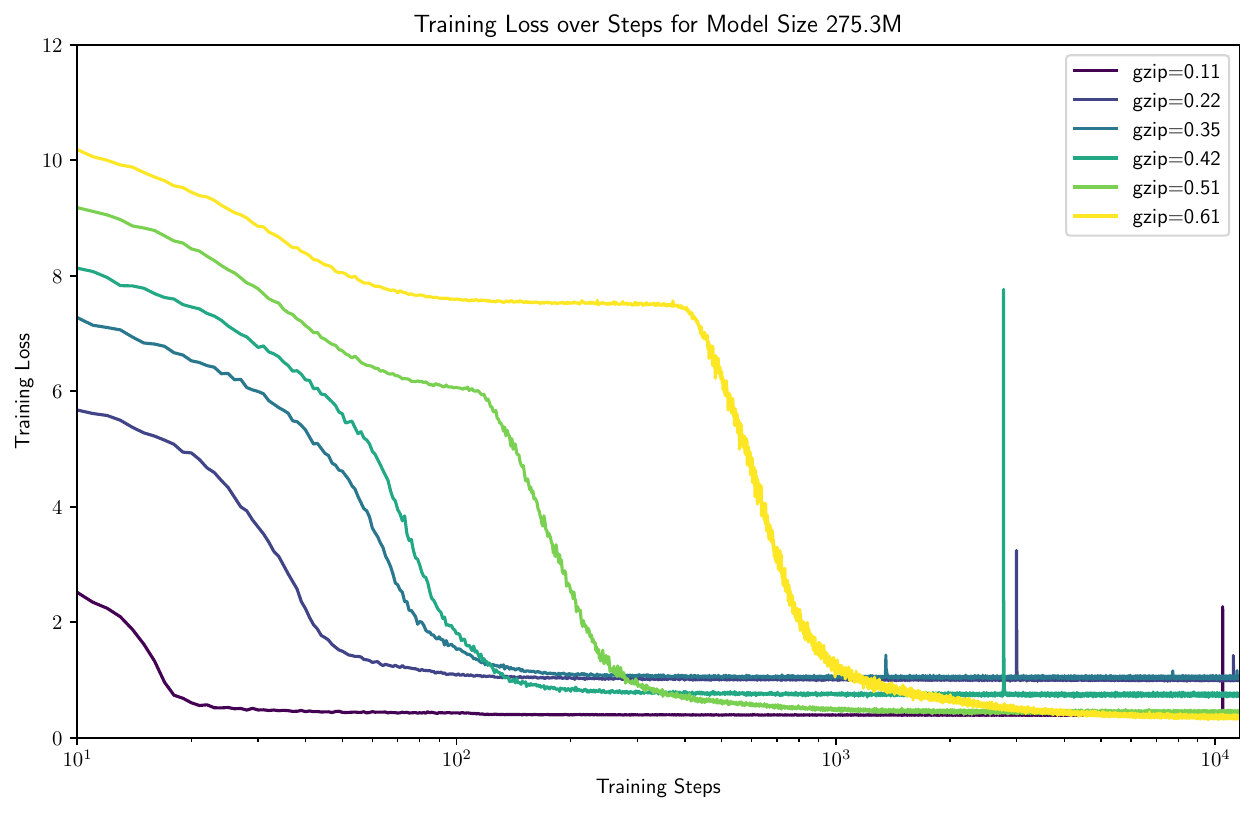}
        \caption{275.3M}
    \end{subfigure}
    \hfill
    \begin{subfigure}{0.32\textwidth}
        \includegraphics[width=\linewidth]{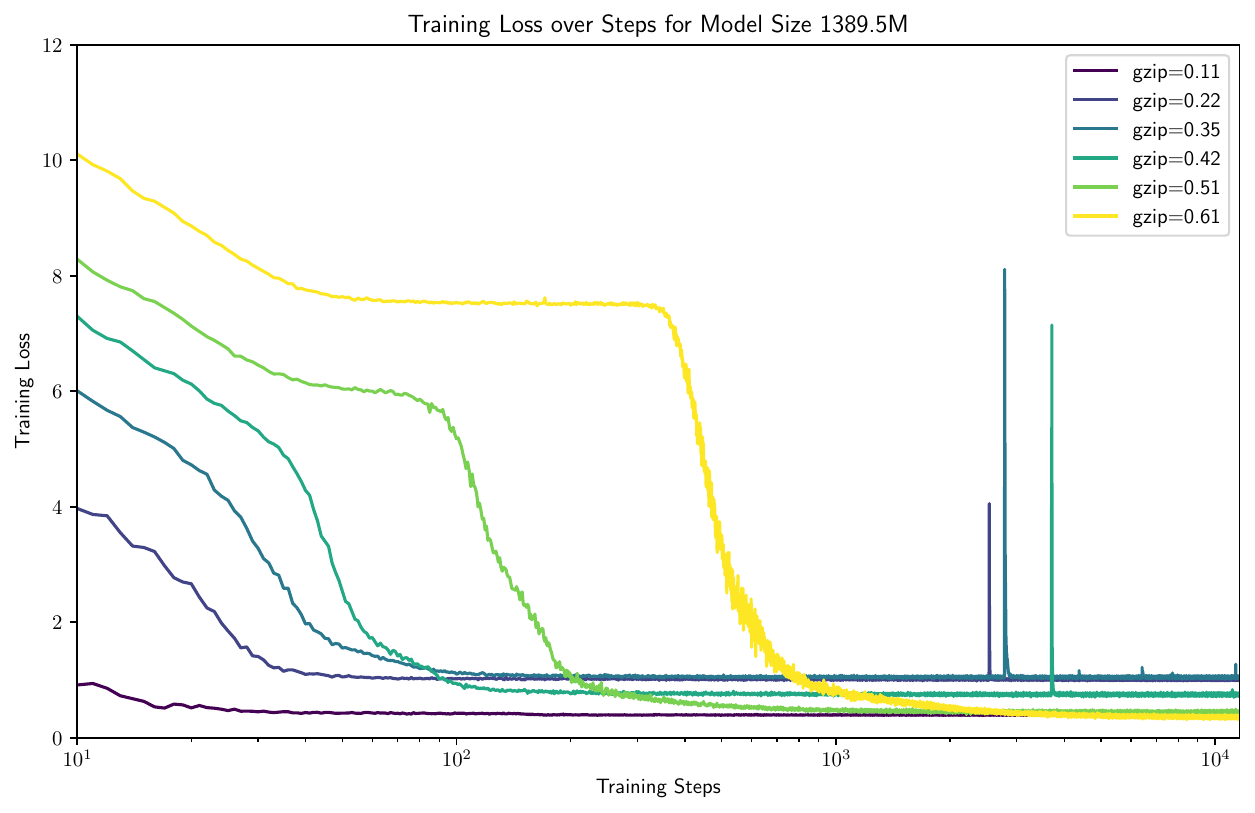}
        \caption{1.4B}
    \end{subfigure}
    \caption{Training curves for models of various sizes trained on 100M token datasets with \texttt{gzip}-compressibilities of 0.11, 0.22, 0.35, 0.42, 0.51, and 0.61. The harder the data is to compress, the longer it takes to train to convergence.}
    \label{training_curves}
\end{figure}

As intuitively expected, the more compressible a dataset is (lower compressibility ratio), the faster that models regardless of size will converge (Fig. \ref{training_curves}). While this shows that we need more compute to model more complex datasets (which is in and of itself notable), we need more evidence to determine if the compute-optimal frontier directionally shifts based on data complexity.
To establish such a non-trivial sensitivity of scaling laws to data complexity, we need to compute the law for each dataset and examine its fitted parameters.

\subsection{Computing Data-sensitive Scaling Laws from \texttt{gzip}-compressibility}

The scaling law functional form proposed by \citet{hoffmann2022training} predicts training loss as a function of model \& dataset size:

\begin{equation}
\label{eq:scaling_law}
L(N, D) = E + \dfrac{A}{N^\alpha} + \dfrac{B}{D^\beta}
\end{equation}

where $N$ is the model's parameter count and $D$ is the training dataset's token count. They claim that $E$ captures the ``entropy of natural text'' (their Sec. 3.3) and that scaling laws are ``independent of dataset'' (their App. C). However, when we fit this function on the training results for each of our PCFG datasets (leveraging the helpful open-source implementation of \citet{muennighoff2024scaling}), we find considerably different laws for each dataset (Tab. \ref{tab:scaling_law_params}).

\begin{table}[ht]
\centering
\caption{Fitted scaling law parameters for each dataset with \texttt{gzip} compressibility}
\begin{tabular}{cccccc}
\toprule
\texttt{gzip} Compressibility & A & B & E & $\alpha$ & $\beta$ \\
\midrule
0.12 & 18.205 & 15.803 & -0.734 & 1.053 & 1.307 \\
0.23 & 15.553 & 13.912 & -0.262 & 0.881 & 1.064 \\
0.32 & 16.236 & 8.829 & -1.446 & 0.930 & 0.597 \\
0.45 & 15.506 & 9.002 & -1693.171 & 0.882 & 0.596 \\
0.60 & 9.061 & 3.525 & 1.174 & 0.559 & 0.140 \\

\bottomrule
\end{tabular}
\label{tab:scaling_law_params}
\end{table}

\begin{figure}[ht]
    \centering
    
    \begin{subfigure}{0.49\textwidth}
        \includegraphics[width=\linewidth]{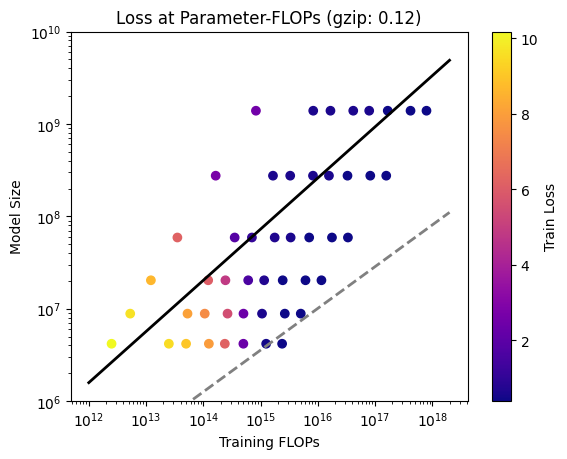}
        \caption{\texttt{gzip}-compressibility = 0.12}
    \end{subfigure}
    \hfill
    \begin{subfigure}{0.49\textwidth}
        \includegraphics[width=\linewidth]{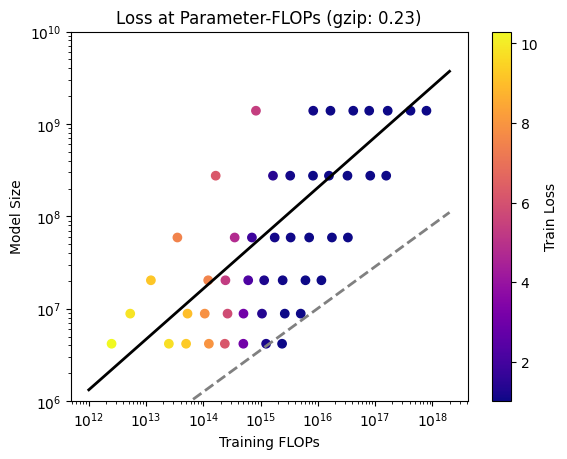}
        \caption{\texttt{gzip}-compressibility = 0.23}
    \end{subfigure}

    \begin{subfigure}{0.49\textwidth}
        \includegraphics[width=\linewidth]{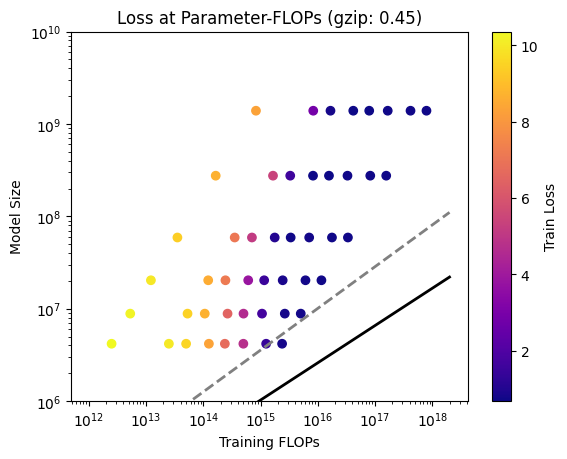}
        \caption{\texttt{gzip}-compressibility = 0.45}
    \end{subfigure}
    \hfill
    \begin{subfigure}{0.49\textwidth}
        \includegraphics[width=\linewidth]{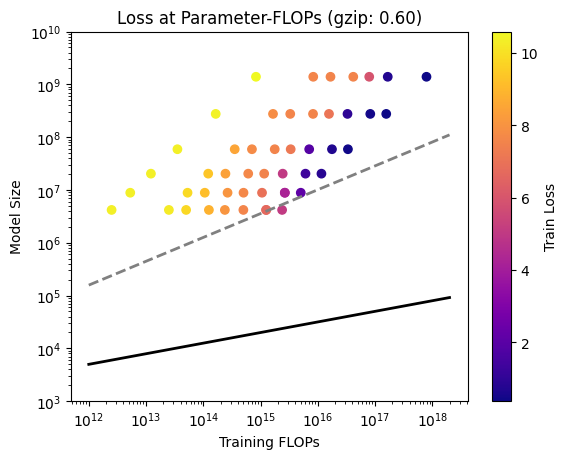}
        \caption{\texttt{gzip}-compressibility = 0.60}
    \end{subfigure}

    \caption{The stationary grey dashed line is the compute-optimal frontier (Eq. \ref{eq:optimal_frontier}) of the allegedly data-agnostic Chinchilla scaling law, and the black line is the same frontier computed from our fitted scaling law for each PCFG dataset. Points are individual training runs colored by their loss. As the data becomes more difficult to compress, the scaling law shifts from being more parameter-preferent (a) than Chinchilla to far more data-preferent (d). Compare Appendix Fig. \ref{fig:scaling_contours} for parameter-data plot.}
    \label{fig:compute_frontiers}
\end{figure}

The scaling law induces a compute-optimal frontier for parameter count derived by \citet{kaplan2020scaling} \& \citet{hoffmann2022training} and simplifiable as:

\begin{equation}
N_{opt}(C) = \left(\dfrac{\alpha A}{\beta B} \left(\dfrac{C}{6}\right)^\beta\right)^{\dfrac{1}{\alpha + \beta}}
\label{eq:optimal_frontier}
\end{equation}

where $C$ is a compute budget in FLOPs. We plot this compute-optimal frontier for Chinchilla as well as our fitted law for each PCFG dataset in Fig. \ref{fig:compute_frontiers}. From a cursory glance, we can see that the frontier of our fitted law progressively becomes more data-preferent as data becomes harder to compress, crossing over Chinchilla's 1-to-1 frontier at some point $0.23 < \texttt{gzip}\text{-compressibility} < 0.45$.

In order to predict scaling law parameters from a dataset's compressibility, let us fit a simple linear regression on the fitted scaling law parameters for each dataset. Recall from Sec. \ref{gzip_measures} that we compute the compressibility $H$ of a dataset $\boldsymbol{D}$ by taking the average ratio of compressed bits to original bits for each element $d$:

\begin{equation}
H(\boldsymbol{D}) = \dfrac{1}{||\boldsymbol{D}||} \sum_{d \in \boldsymbol{D}} \dfrac{||\texttt{gzip}(d)||}{||d||} 
\end{equation}

\begin{figure}[ht]
    \centering
    
    \begin{subfigure}{0.49\textwidth}
        \includegraphics[width=\linewidth]{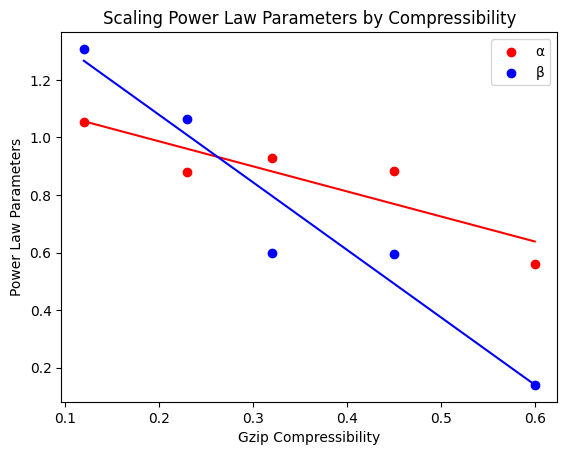}
    \end{subfigure}
    \hfill
    \begin{subfigure}{0.49\textwidth}
        \includegraphics[width=\linewidth]{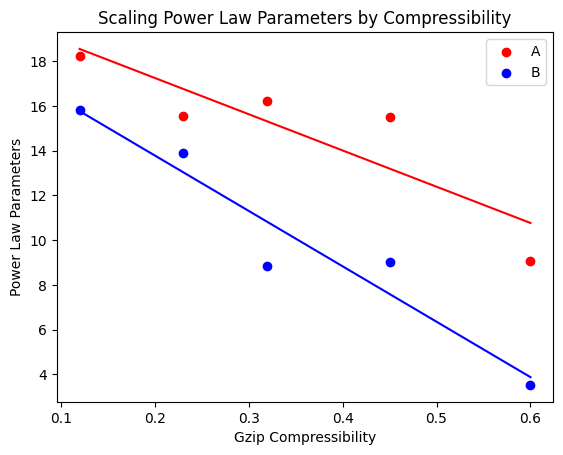}
    \end{subfigure}

    \caption{As data becomes harder to compress, the scaling law's fitted parameters trend downwards at different rates ($p < 0.05$), with $\alpha$ and $\beta$ intersecting around 0.27—the compressibility point at which parameters \& data should be scaled roughly equiproportionally.}
    \label{fig:gzip_params}
\end{figure}

So once we've fit lines to predict each parameter (E, A, B, $\alpha$, $\beta$) from $H$, we can re-define each parameter as a function of compressibility:

\begin{equation}
\forall x \in \{E, A, B, \alpha, \beta\}: x(H) = m_x H + n_x
\end{equation}

where $m_x$ and $n_x$ are just our fitted linear regressions' parameters. These fitted values (along with the regressions' $p$-values) are presented in Tab. \ref{tab:gzip_law_params} and the linear regressions are visualized in Fig. \ref{fig:gzip_params}. They're all pretty much monotonically decreasing at different rates with an interesting $\alpha$-$\beta$ intercept at $H \approx 0.27$. It's also interesting to note that $E$, originally set as the constant `entropy of natural language', is the only parameter that loosely increases with $H$ (though not significantly)—our proxy for complexity or entropy.

\begin{table}[h]
    \centering
    \caption{Fitted values for $x(H)$ that make the scaling law parameters ($x \in \{E, A, B, \alpha, \beta\}$) a linear function of data compressibility ($H$).}
    \begin{tabular}{c|ccc}
    $x$ & $m_x$ & $n_x$ & $p_x$ \\
    \hline
    E & 3.92 & -1.56 & .272 \\
    A & -16.20 & 20.48 & .048 \\
    B & -24.77 & 18.73 & .009 \\
    $\alpha$ & -0.87 & 1.16 & .043 \\
    $\beta$ & -2.34 & 1.55 & .008 \\
    \end{tabular}
    \label{tab:gzip_law_params}
\end{table}

Now, we can reparameterize Eq. \ref{eq:scaling_law} as a function of compressibility $H$:

\begin{equation}
\label{eq:gzip_law}
L(N, D, H) = E(H) + \dfrac{A(H)}{N^{\alpha(H)}} + \dfrac{B(H)}{D^{\beta(H)}}
\end{equation}

However, since our experiments are considerably smaller scale and primarily on PCFG data, we also present a form of our data-dependent scaling law as an adjustment of Chinchilla, where $\varepsilon$ is the weightage of our adjustment for the \texttt{gzip}-compressibility of the training data (and prime ($'$) parameters are the Chinchilla constants).

\begin{equation}
\label{eq:gzip_law_adjusted}
L(N, D, H) = (1 - \varepsilon) E' + \varepsilon E(H) + \dfrac{(1 - \varepsilon) A' + \varepsilon A(H)}{N^{(1 - \varepsilon) \alpha' + \varepsilon \alpha(H)}} + \dfrac{(1 - \varepsilon) B' + \varepsilon B(H)}{D^{(1 - \varepsilon) \beta' + \varepsilon \beta(H)}}
\end{equation}

\subsection{Eliminating Syntactic Parameters as a Confounder of Compressibility}

\begin{figure}[ht]
    \centering
    
    \begin{subfigure}{0.49\textwidth}
        \includegraphics[width=\linewidth]{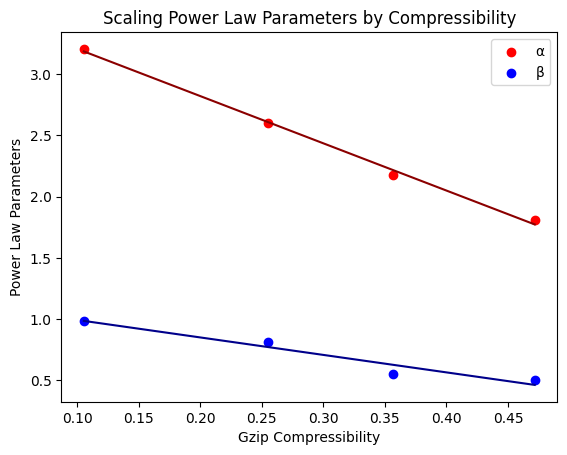}
    \end{subfigure}
    \hfill
    \begin{subfigure}{0.49\textwidth}
        \includegraphics[width=\linewidth]{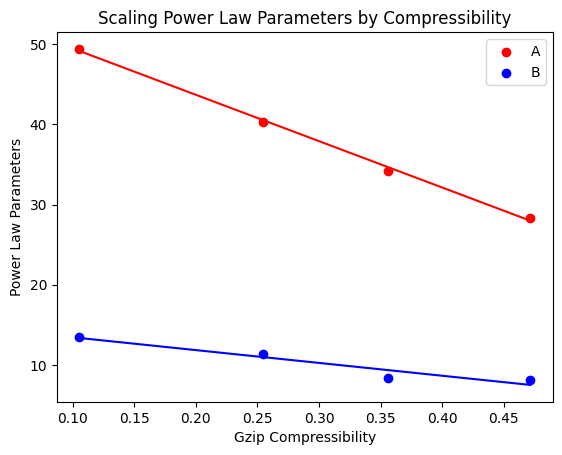}
    \end{subfigure}

    \caption{When holding terminal count (vocab size) the same and varying other syntactic properties, scaling laws still shift across \texttt{gzip}-compressibility, with even stronger correlation.}
    \label{fig:isovocab_params}
\end{figure}

Given just the above experiments, we have not addressed the possibility that our compressibility measure is confounded by some underlying syntactic property (e.g. vocab size). To address this concern, in Fig. \ref{fig:isovocab_params} we provide results showing that when holding vocab size steady and changing other syntactic properties (Tab. \ref{tab:isovocab_parameters}), \texttt{gzip}-compressibility still predicts scaling law parameter shifts (with an even stronger correlation than in the increasing vocab size setting). 

\begin{table}[ht]
\centering
\caption{Syntactic parameters for PCFG datasets holding terminal count the same}
\begin{tabular}{ccccc}
\toprule
Compressibility & Nonterminals & Terminals & RHS Options & RHS Length \\
\midrule
0.11 & 3 & 300 & 2 & 2 \\
0.25 & 10 & 300 & 5 & 3 \\
0.36 & 20 & 300 & 10 & 5 \\
0.47 & 50 & 300 & 20 & 10 \\
\bottomrule
\end{tabular}
\label{tab:isovocab_parameters}
\end{table}

\begin{figure}[ht]
    \centering
    
    \begin{subfigure}{0.49\textwidth}
        \includegraphics[width=\linewidth]{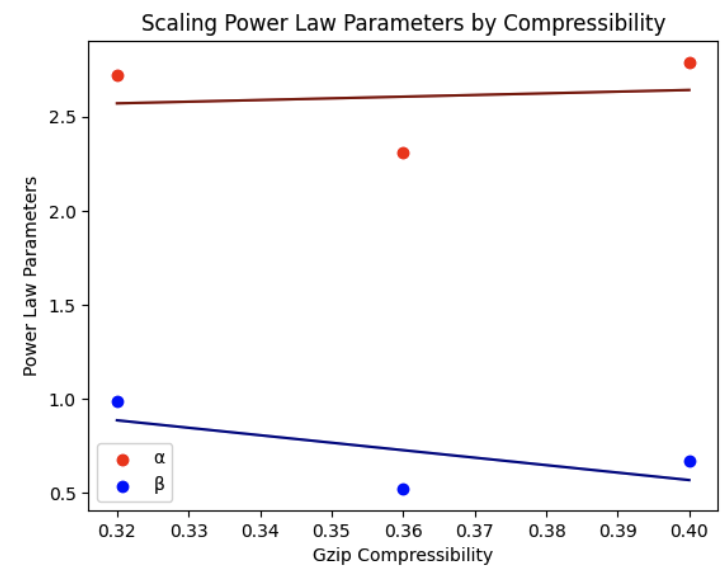}
    \end{subfigure}
    \hfill
    \begin{subfigure}{0.49\textwidth}
        \includegraphics[width=\linewidth]{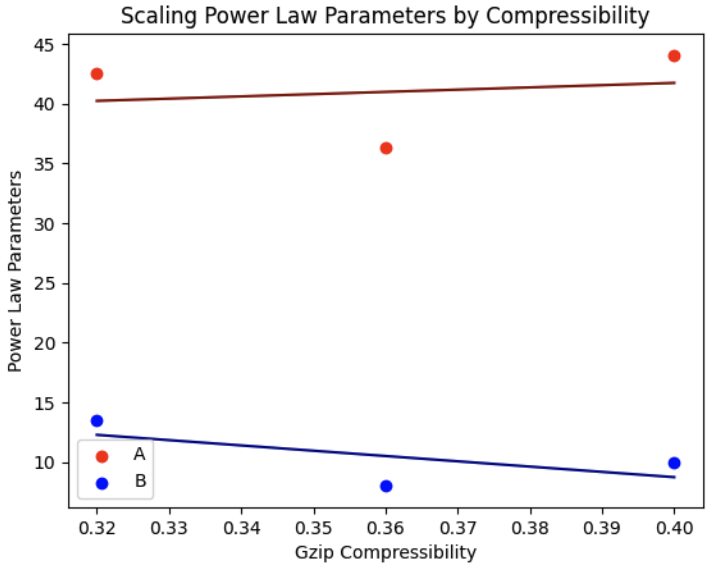}
    \end{subfigure}

    \caption{When non-monotonically varying syntactic parameters but maintaining roughly the same \texttt{gzip}-compressibility (0.32 - 0.40), scaling law parameters (especially $\alpha$ and $A$) do not meaningfully shift.}
    \label{fig:isogzip_params}
\end{figure}

We also empirically show the contrapositive in Fig. \ref{fig:isogzip_params}, demonstrating that when we widely vary syntactic properties (Tab. \ref{tab:isogzip_parameters}) but such that the datasets' final \texttt{gzip}-compressibility are all the same, there is no significant shift in scaling law parameters.

\begin{table}[ht]
\centering
\caption{Syntactic parameters for holding \texttt{gzip}-compressibility the same}
\begin{tabular}{ccccc}
\toprule
\texttt{gzip}-Compressibility & Nonterminals & Terminals & RHS Options & RHS Length \\
\midrule
0.32 & 10 & 500 & 5 & 10 \\
0.36 & 20 & 300 & 10 & 5 \\
0.40 & 30 & 200 & 15 & 20 \\
\bottomrule
\end{tabular}
\label{tab:isogzip_parameters}
\end{table}

Although the intersection behavior observed in Fig. \ref{fig:gzip_params} is not observed in the iso-vocab case (Fig. \ref{fig:isovocab_params}), the steeper negative slope of $\alpha$ v.s. $\beta$ (and $A$ v.s. $B$) implies the same phenomenon of increased data-preference as \texttt{gzip}-compressibility grows.

Thus we have shown that scaling laws are dependent on training data and \texttt{gzip}-compressibility is a strong predictor of how data complexity will impact scaling properties. We provide fitted coefficients (Tab. \ref{tab:gzip_law_params}) for predicting scaling parameters as a function of $H$ (\texttt{gzip}-compressibility), and offer an adjustment-based formulation (Eq. \ref{eq:gzip_law_adjusted}) of the data-dependent scaling law to make our small scale synthetic experiments more adaptable to training frontier models.

\section{Discussion}

We have shown that the Chinchilla scaling law \citep{hoffmann2022training} and similar laws that claim to be data-agnostic are in fact web-text-specific instances of a broader family of scaling laws sensitive to information-theoretic measures of training data. Not only do datasets that are difficult to compress (regardless of syntactic specifics) require more compute, they also require a different trade-off between model size \& data.

Why does a dataset with a \textbf{certain level of compressibility $H$ result in the specific scaling parameters} that it does? And why does increased \texttt{gzip}-compressibility result in increased preference for data over parameter count? These questions warrant further exploration from a theoretical perspective—we hope to leverage connections between syntax \& information theory to answer this question in future work.

On the empirical side, an especially promising avenue is in LM's for code since code datasets have significantly lower \texttt{gzip}-compressibility than natural language. Therefore, we can expect the \textbf{compute-optimal scaling law for code to have stronger parameter-preference} than Chinchilla. We are currently running experiments at 1.92e19 FLOPs (some \texttt{gzip}-compressibility analysis in Fig. \ref{fig:code_compressibility} and preliminary runs in Fig. \ref{fig:code_learning}) to compare performance based on Chinchilla's allocation and our data-adjusted scaling law's (Eq. \ref{eq:gzip_law_adjusted}) allocation. As highlighted in Sec. \ref{intro}, scaling up this finding could save \$278,000 in H100 hours when training a single relatively small (6B parameter) code-generation model.

Beyond a deeper theoretical understanding of scaling laws \& more optimal compute allocation for code-generation LMs, our findings also open possibilities for using compressibility as a metric in training data filtering, curriculum learning ordering, and fine-tuning data requirement estimation. Anecdotally, we have heard from researchers at frontier AI labs that compression algorithms are already used in production for filtering LLM training data.

The call-to-action here is not per se to use our data-dependent scaling law to decide how to spend your next million dollars of compute, but rather to begin investigating whether performance discrepancies between models trained on different datasets are explainable by simple information-theoretic measures (e.g. \texttt{gzip}-compressibility) of training data. If so (as experienced by \citet{bi2024deepseek}), it may be worth building systems to dynamically determine compute allocation for each training dataset, with the general rule of thumb that easy-to-compress datasets prefer parameters and hard-to-compress datasets prefer data.

\section{Limitations}

The work presented here is restricted to the synthetic PCFG setting and does not yet explicitly show generalization to predicting different scaling properties of real-world datasets. Our method of modulating training data complexity by varying syntactic properties of a PCFG, though naturalistic, is fairly specific and may still be prone to confounders. In addition, there are features of training data that affect scaling but appear to not be modeled well by \texttt{gzip}-compressibility. Alternative compression algorithms \& information-theoretic measures of training data were not tested though they likely all broadly correlate with \texttt{gzip}-compressibility.

\section{Acknowledgements}
We are grateful to Alon Albalak, Rylan Schaeffer, Charlie Snell, Ethan Caballero, Martin Ziqiao Ma, Sankeerth Rao Karingula, Guillaume Lample, Vibhu Sapra, Aryaman Arora, Tristan Thrush, and several anonymous Twitter users for their thorough reviews \& feedback on drafts of this paper.

This work was also sponsored by generous compute grants from Luke Piette (RunPod), Kye Gomez (Swarms), Vincent Weisser \& Johannes Hagemann (Prime Intellect), and of course the rest of my team at Reworkd—Asim Shrestha \& Adam Watkins.

\bibliography{citations}

\clearpage
\appendix

\section{Appendix}

\begin{table}[ht]

\centering
\caption{Architectural properties for each model size}
\begin{tabular}{cccccc}
\toprule
Parameter Count & Hidden Size & Intermediate Size & Hidden Layers & Attention Heads \\
\midrule
4.2M & 64   & 128   & 2  & 1  \\
8.8M & 128  & 256   & 4  & 2  \\
20.3M & 256  & 512   & 6  & 4  \\
59.0M & 512  & 1024  & 10 & 8  \\
275.3M & 1024 & 2048  & 20 & 16 \\
1.4B & 2048 & 4096  & 30 & 32 \\
\bottomrule
\end{tabular}
\label{tab:architecture}
\end{table}

\begin{figure}[ht]
    \centering
    
    \begin{subfigure}{0.49\textwidth}
        \includegraphics[width=\linewidth]{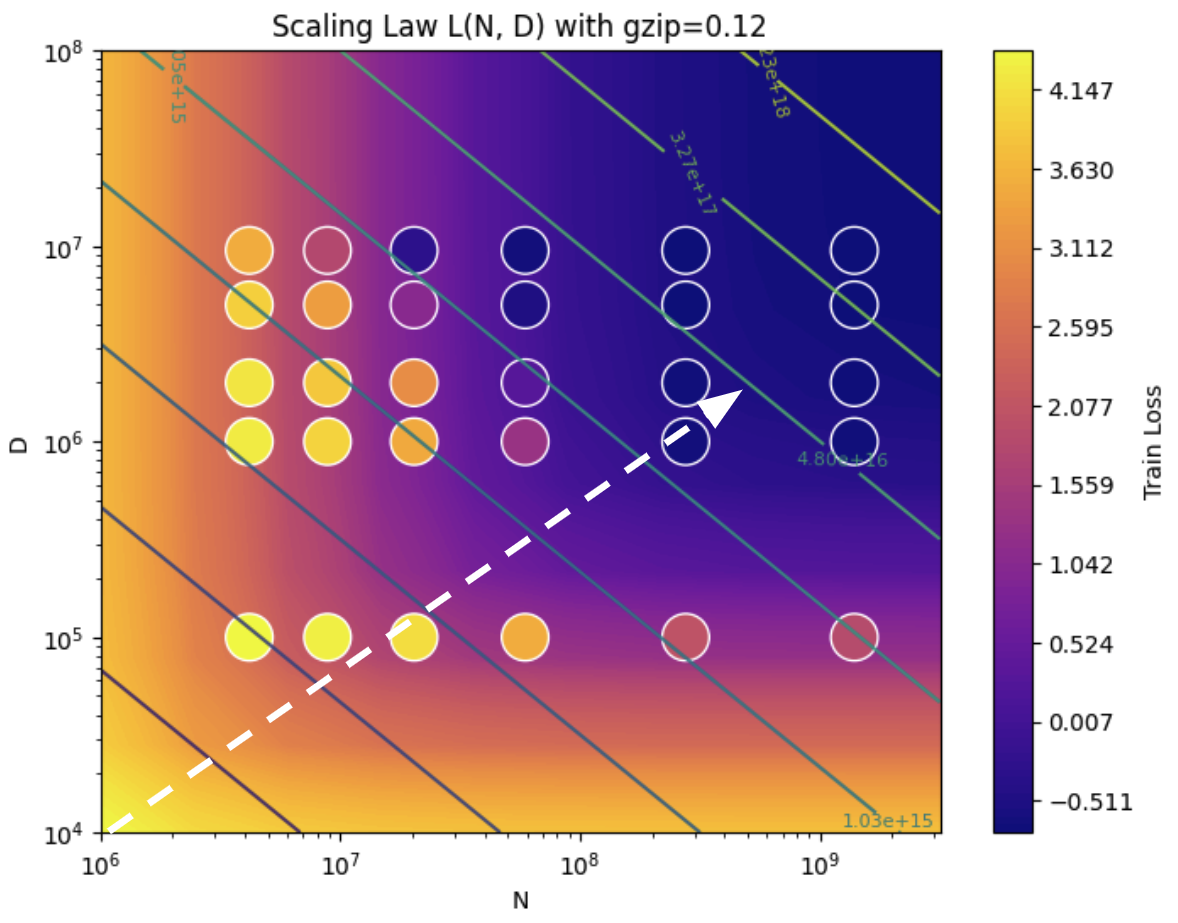}
    \end{subfigure}
    \hfill
    \begin{subfigure}{0.49\textwidth}
        \includegraphics[width=\linewidth]{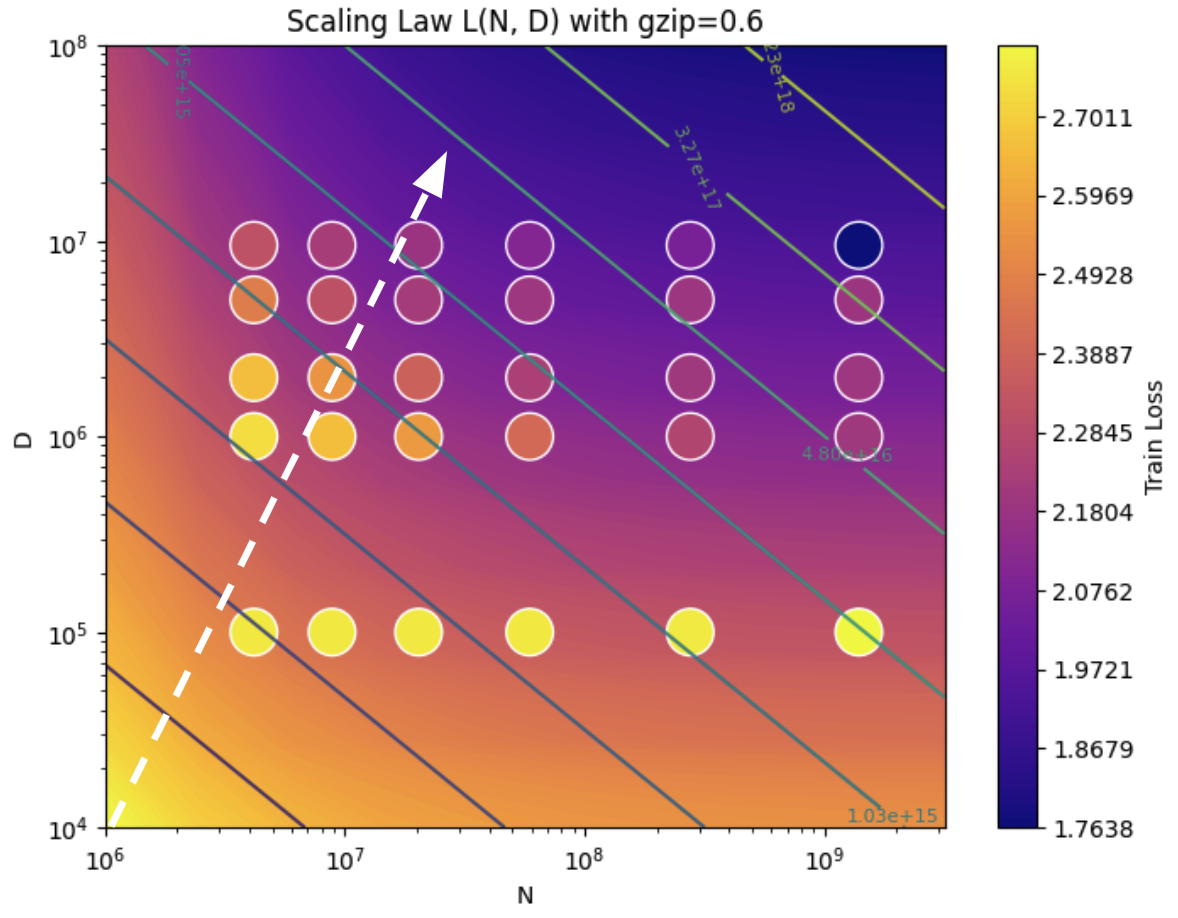}
    \end{subfigure}

    \caption{Scaling contours in parameter-data space, showing how for dataset with higher \texttt{gzip}-compressibility (0.60), not only is more compute required, but the optimal frontier increasingly prefers data.}
    \label{fig:scaling_contours}
\end{figure}

\begin{figure}[ht]
    \centering
    \includegraphics[width=\linewidth]{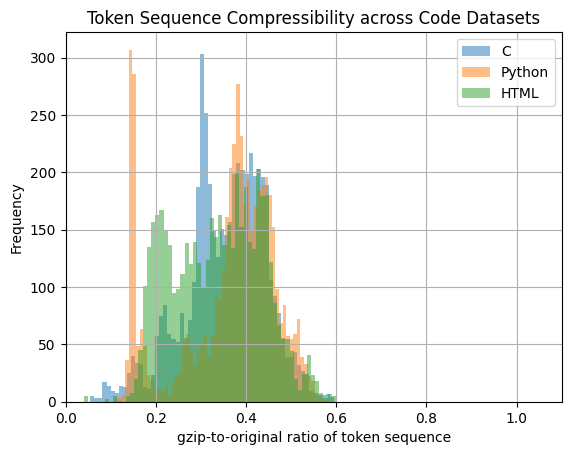}

    \caption{Real-world code datasets of different languages have slightly different compressibilities, but all considerably less than natural language.}
    \label{fig:code_compressibility}
\end{figure}

\begin{figure}[ht]
    \centering
    
    \begin{subfigure}{0.49\textwidth}
        \includegraphics[width=\linewidth]{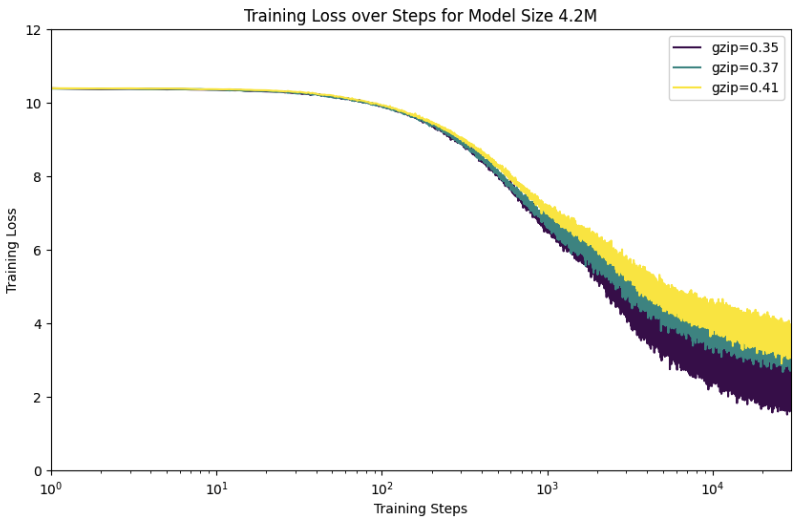}
    \end{subfigure}
    \hfill
    \begin{subfigure}{0.49\textwidth}
        \includegraphics[width=\linewidth]{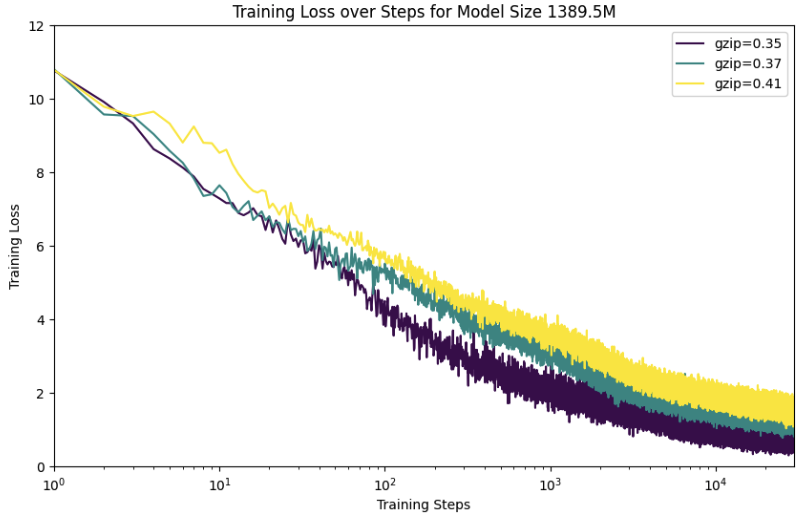}
    \end{subfigure}

    \caption{In real-world code datasets with even slightly different compressibilities (HTML, C, \& Python), slightly more compute is required (regardless of model size) as data becomes harder to compress. Further work is required to compare their scaling laws in relation to natural language.}
    \label{fig:code_learning}
\end{figure}

\end{document}